# Robust Dialogue Understanding in HERALD

## Vincenzo PALLOTTA and Afzal BALLIM

LITH-MEDIA laboratory

Swiss Federal Institute of Technology - Lausanne

IN F Ecublens

1015 Lausanne, Switzerland,

{Vincenzo.Pallotta,Afzal.Ballim}@epfl.ch

## Abstract

We tackle the problem of robust dialogue processing from the perspective of language engineering. We propose an agent-oriented architecture that allows us a flexible way of composing robust processors. Our approach is based on Shoham's Agent Oriented Programming (AOP) paradigm. We will show how the AOP agent model can be enriched with special features and components that allow us to deal with classical problems of dialogue understanding.

## 1 Introduction

The design of robust Dialogue Systems (DS) is nowadays one of the most challenging issues in NLP. As remarked in (Allen et al., '01) the role played by the software infrastructure is a non-trivial one. In the HERALD[1] architecture (Ballim et al., '00b), we considered a very general programming paradigm that incorporates most of the desired features for the rational design of "intelligent" systems. In Dialogue Management Systems, intelligence relies on the efficient combination of both reactive and rational behaviour (i.e. decision making). Moreover, such systems should be able to deal with often-unforeseen situations.

### 1.1 Language Engineering and distributed NLP

It is apparent that Software Engineering (SE) and in particular Natural Language Engineering (NLE) needs to consider both theoretical and practical issues when adopting a design methodology. Moreover, a specification language needs to have its sound implementation counterpart. For instance, the Natural Language Engineering architecture GATE[2] (Cunningham et al., '95), although very useful for designing modular NLP systems, doesn't seem suitable for implementing a dialogue system[3], because of a design that is optimised for information extraction: rigid module coupling and document transformation-based communication. Nevertheless, there is an ever-increasing interest towards distributed cooperative approaches to NLP. On the two extremes of these approaches, we have unsupervised and supervised coordination of intervening autonomous modules. For instance, the TREVI toolkit (Basili et al., '00) provides an environment for the rational design of object-oriented distributed NLP application where the cooperation between modules is statically decided but dynamically coordinated by a dataflow-based object manager. In contrast, the incremental Discourse Parser Architecture (DPA) (Cristea, '00) assumes no predefined coordination schema but rather the spontaneous cooperation of well-defined autonomous linguistic experts.

The above approaches to the design of NLP systems are all motivated by the need of modelling the system behaviour by means of *content information* rather than by means of exclusively general principles. The coexistence of multiple theories may allow the system to select the most appropriate analysis strategy and heuristically set different *tuning parameters*[4] that may radically alter their performance and outcomes.

In order to cope with mixed-initiative dialogue we considered the possibility of building our system by adopting a distributed composition strategy. We adopted some results obtained in NLP systems like TRIPS (Allen et al., '00), TREVI (Basili et al., '00) and DPA (Cristea, '00). In addition, we looked for a design environment providing a set of software engineering tools for the rational design of NLP applications.

### 1.2 Modularity in Linguistics

We found inspiration for the development of our vision from recent trends in theoretical linguistics where modularity is being taken into greater consideration for language and discourse analysis of written text. In (Nolke & Adam, '99) some principles and methodological issues seem to provide us an underlying theoretical framework: instead of postulating a rigid decomposition of language into linguistic components (e.g. syntax, semantics, pragmatics), they propose to apply modularity to analysis by defining autonomous analysis modules, which in turn may consider different aspects of the language. The decomposition in linguistic levels may be still useful but it should not be considered as a pre-theoretical hypothesis on the nature of the language. Instead, analysis modules may take advantage of this decomposition to identify what features of the language should be considered. Modular independence (i.e. autonomy) is also a key issue. In order to provide a global modelling of the language, autonomous analysis modules can be composed by means of global meta-rules. To achieve this goal, modules should have a common interchange language and provide a global accessibility. The main advantages to methodological modularity in linguistics can be summarized in the following points:

- Extraction of language features at different granularities (i.e. phonemes, words, phrases, discourse, etc.).

- Co-existence of different analysis strategies that embody different principles.

- Coverage obtained as the composition of different (and possibly overlapping) language models and analysis techniques.

---

[1] Hybrid Environment for Robust Analysis of Language Data.

[2] General Architecture for Text Engineering

[3] The new release of GATE will introduce agent-based language processing capabilities and a new approach to linguistic resources distribution.

[4] Note that in Natural Language Processing the choice of a parameter could be the selection of a suitable linguistic resource (or the selection of its subpart).

While the main goals in linguistics are the description of the language and the explanation of language phenomena, computational linguistics is more interested on how the above these methodological improvements may help in the design of NLP applications. Our vision can be outlined as follows.

- We believe that by eliminating the rigid decomposition in linguistic levels we can move from sequential composition to distributed composition of processing modules that can take into account more than a single linguistic level at time.

- We can complete the analysis of "ideal" language driven by principles with the processing of "real" language driven by heuristics. The combination of the theoretical and the practical account to the language processing will necessarily lead to more robust NLP.

### 1.3 Previous work

We have worked on robust parsing in the context of the project ROTA[5] (Ballim & Pallotta, '99), (Ballim & Pallotta, '00), which showed that a logic-programming framework for NLP based on Definite Clause Grammars (DCG) can be fruitfully extended for robust parsing (Ballim & Russell, '94). In this project, we achieved some improvements of the robust parser LHIP[6] (Lieske & Ballim, '98). The underlying idea of LHIP is very attractive since it allows one to perform parsing at different arbitrary levels of "shallowness". It can be envisioned that its role in a dialogue system would be in:

- Chunk extraction
- Implementation of semantic grammars
- Concept spotting

---

[5] ROTA stands for Robust Text Analysis.

[6] LHIP stands for Left-corner Head-driven Island Parser.

- Extraction of dialogue acts

LHIP has been fruitfully used within another past project: ISIS[7] (Chappelier et al., '99), (Ballim et al., '00a), where we experimented with an approach to Question & Answering Systems combining stochastic parsing and robust semantic parsing. We have shown that a fixed composition strategy of these techniques is adequate for a certain class of fixed initiative spoken dialogue system.

## 2 The HERALD architecture

When switching to mixed-initiative dialogue system a more natural form of interaction is required: it is crucial to rely on autonomous, loosely coupled interacting components. We believe that this kind of architecture provides the necessary computational background for developing portable and reusable dialogue systems with a clear separation between discourse modelling and task/domain reasoning. We can summarize the HERALD software engineering requirements as follows:

- *Rule based specification of system's modules composition*. Modules are be loosely coupled and should allow dynamic reconfiguration of the system topology. Composition rules should account for the types of data object that modules are supposed to exchange.

- *Dynamic task assignment based on contextual information*. Coordination modules should be able to access information about others modules capabilities, evaluate their performance and select among the best response when multiple modules are activated in parallel to accomplish similar or competing goals.

- *Logic based decision support*. The coordination decisions should be taken rationally.

### 2.1 Agent-oriented Software Engineering in HERALD

In choosing a platform for the implementation of the HERALD prototype, we looked for a suitable design methodology and programming model. The *Mentalistic Agent* model seemed the most appropriate since it allows us to design multi-agent systems where each participant play a different role and manages its own knowledge base that uses for taking rational decisions.

As the starting point for our framework, we considered the Agent Oriented Programming model (AOP) (Shoham, '93). AOP can be viewed as a specialization of object-oriented programming. Objects become agents by redefining both their internal state and their communication protocols in intentional terms. Whereas normal objects contain arbitrary values in their slots and communicate using unstructured messages, AOP agents contain *beliefs*, *commitments*, *choices*, and the like; they communicate with each other via a constrained set of *communicative acts* such as *inform*, *request*, *promise*, and *decline*. The state of an agent is explicitly defined as a *mental state*. AOP does not prevent the construction of BDI (Rao & Georgeff, '91) agents since AOP is an *open programming environment* and does not preclude the adoption of more refined agent models ranging from pure reactivity to full rationality.

---

[7] Interaction through Speech with Information Systems.

AOP agents can be specified by set of *behavioural rules*, which can be viewed as WHEN-IF-THEN triggers whose bodies are allowed to have any combination between the possible actions and mental changes. AOP can be also considered a methodology for designing *open systems*, in which additional processing components can be easily plugged-in.

### 2.1.1 Extensions to AOP in HERALD

We review now some of the features we added to AOP original proposal in order to support the design of NLP applications.

*Mental state management in ViewGen.* The ViewGen system (Wilks & Ballim, '87), (Ballim & Wilks, '90), (Ballim & Wilks, '91a), (Ballim & Wilks, '91b) is intended for use in modelling autonomous interacting agents and it is an implemented version of ViewFinder specifically tailored for modelling agents' mutual beliefs. A belief environment represents each agent's belief space and it may use nested environments to represent other's agent beliefs spaces. As pointed out in (Ballim, '93) the attribution of belief by means of ascription can be generalised to other mental attitudes providing a common theory of mental attitude attribution.

*ViewFinder.* ViewFinder (Ballim, '92) is a framework for manipulating environments. *Environments* (or views, or partitions, or contexts) are aimed at providing an explicit demarcation of information boundaries, methodological benefits (allowing on to think about different knowledge spaces), as well as processing one (allowing for local, limited reasoning, helping to reduce combinatorial problems, etc.). The ViewFinder framework provides the foundations for the following issues related to the manipulation of environments:

- correspondence of concepts across environments
- operations performed on environments
- maintenance of environments.

Relationships between environments can be specified hierarchically or using an explicit mapping of entities. Each environment has an associated axiomatization and a reasoning system.

*Capability Description Language.* A software layer for the processing of the KQML (Labrou & Finin, '97) primitives for agent recruitment in multi agent systems has been implemented. The *capability broker* is the agent who has the knowledge about the capabilities of different "problem-solving" agents who "advertise" their capabilities by means of a *Capability Description Language* (CDL) (Wickler, '99). The utility of a capability brokering mechanism is apparent if our goal is to conceive an open architecture where modules can be plugged-in without any direct intervention.

*Fluent Logic Programming.* We contributed to the design of *Fluent Logic Programming* (FLP) action language and to the implementation of its proof procedure (Pallotta, '99), (Pallotta, '00b), (Pallotta, '00a). The nice feature of FLP is that it allows to model incomplete information about world states, which can be inferred on demand in a particularly efficient way. This capacity is definitely useful to model agents' hypothetical behaviours.

*IRC Facilitator.* We implemented the IRC[8] PAC that allows an agent to post messages on a shared message-board. The IRC PAC allows the construction of agents capable of connecting to an IRC server that offers routing and naming facilities. IRC can be viewed as an alternative agent communication infrastructure based on a hub topology that provides at the same time peer-to-peer and broadcasted message passing.

### 2.1.2 The proposed architecture

In a modular system the choice of what information the modules may exchange is a critical issue, in particular when a module does not know in advance what kind of processing is required on the received data, a situation which is fairly frequent in dialogue systems. We reproduce in our architecture the dataflow-oriented threading mechanism of the TREVI architecture by mapping it into a suitable usage of typed KQML messages. Here the coordinating agent that holds a dynamically constructed view of the system takes the role of the Flow-manager and the flow-graph is represented by a nested environment representation in ViewFinder.

Several agents having a general internal structure compose the HERALD architecture. An *HERALD reasoning agent* encapsulates an inference engine and a local knowledge base, as well as a core set of rule devoted to the management of its mental state and to the interface with the coordination agent (i.e. the *Knowledge Mediator Agent*). Special purpose agents are included for the *multi-modal input management* and for specific pre-processing of the input (i.e. an *Interpretation Manager*). Figure 1 describes the HERALD architecture.

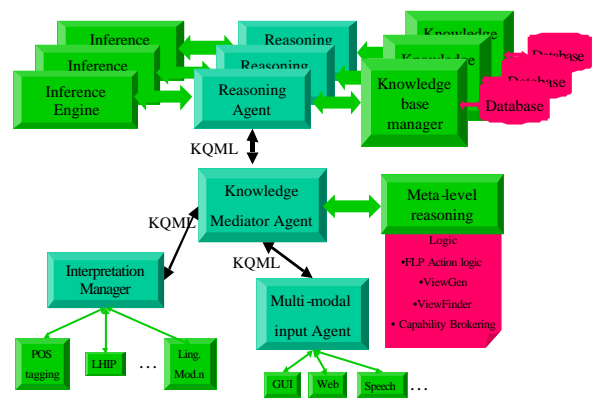

Figure 1: The HERALD architecture

---

[8] IRC stands for International Relay Chat. IRC is a widely used Internet chat server.

# 3 Dialogue understanding in HERALD

In this section, we will outline how is it possible to import design principles from the state-of-the-art dialogue systems into the HERALD architecture. We focus only on few relevant aspects and where left unspecified we adopt the same principles as those proposed in (Allen et al., '01). We consider in this work the Information State Approach to Dialogue Modelling (ISDMS) (Larsson & Traum, '00) and we show how a Dialogue Manager based on this approach can be reconstructed and efficiently implemented as an AOP agency.

## 3.1 Incremental robust interpretation

The need of a robust treatment of users input is amplified by intrinsic recognition errors induced by Automatic Speech Recognisers (ASR). A viable approach is to combine different levels of shallowness in the linguistic analysis of utterances and produce sets of ranked utterance's interpretations. Competing analyses may be compared with respect to the confidence levels we assign to the producing linguistic modules (or the macro-modules that encapsulate some complex processing). We consider the following robust analysis components to be encapsulated by AOP agents:

- <u>Partial and adaptive parsing</u>. Instead of enlarging the set of generative rules to cope with extra-grammatical phenomena, our robust parser LHIP allows us to approximate the interpretation adopting some heuristics in order to gather sparse (possibly correct) analyses of sub-constituents. Heuristics can be also used to automatically determine what domain-related features of the language consider and thus to select an appropriate sub-model.

- <u>Underspecified semantics</u>. Studies have shown that in some cases it is necessary to delay the decision of how to build a semantic interpretation up to the moment when additional contextual information is obtained. Once we are able to construct a logical form for a given utterance, (e.g. by possibly exploiting already accumulated additional information) it could be the case that spurious ambiguity remains which can be solved only at later stages of the analysis. We adopted some methods for representing semantic ambiguity (e.g. multiple quantifiers bindings) in an efficient and compact way (Poesio, '95).

- <u>Abductive discourse interpretation</u>. Interpretation problems like reference resolution, interpretation of nominal compounds, the resolution of syntactic ambiguity and metonymy and schema recognition require an adaptive inference at the level of semantic logical form (Hobbs et al., '93).

The main problem with a system that allows open interpretations is that of the combinatorial explosion of multiple interpretations. The worst case is when we sequentially combine modules producing multiple interpretations without any selection mechanism that reduces the search space and keeps the computation tractable. Unfortunately, ambiguities arisen at a certain linguistic level could be only solved with the information provided at higher levels. If we do not consider an incremental approach we are not able to exploit this information: a transformational approach would force us to generate all the interpretation of one level and feed the subsequent linguistic module with all the produced interpretations. In the best case, we are only able to rank the hypotheses and select the n-best.

Once the utterance has been interpreted (both multiple or underspecified), the Interpretation Manager is able to generate the communicative acts as KQML messages. Only plausible recognized communicative acts are "assimilated" in Discourse Context, meaning that contextual interpretation and pragmatic reasoning is required in order to resolve ellipses and anaphora.

## 3.2 Information State and Dialogue Moves

AOP reflects the principles of the Information State approach to dialogue modelling proposed by (Larsson & Traum, '00). *Dialogue moves* can be coded in fact by means of KQML performatives whereas *update rules* are special cases of *behavioural rules*. AOP poses no restriction on how to structure agent's mental state except for the fact that the formalism allows to express conditions on the represented objects.

There is currently no support for implementing rule selection strategy in AOP. Rules are applied following a kind of "don't care" non-determinism in the style of committed choice in constraint logic programming[9]. A future interesting extension would be certainly to provide a more flexible rule selection mechanisms.

## 3.3 Mental state recognition

ViewGen can be used for plan recognition from speech-acts by a suitable integration of planning, ascription and inference (Lee, '98), (Lee & Wilks, '96), (Lee, '97). Lee proposes to overcome the limitations of the original implementation of ViewGen following the theoretical foundations and the generalisation of ViewFinder, extending the representational framework to cope with agents' mental attitudes by means of typed environments. The type considered by Lee are those of interest in the case of plan recognition from dialogue (e.g. goals and intentions). Based on this extension he proposes the amalgamation[10] of ViewGen with the Partial Order Clausal Link (POCL) planner (McAllester & Rosenblitt, '91). His work led to the successful treatment of a set of speech acts partially based on the Bunt's taxonomy (Bunt, '89) and empirically tested on a dialogue corpus. The notion of agent stereotype is extended to *situations types* (i.e. dialogue types or protocols) and *discourse types* (triggered by the actual linguistic context [11]). Finally Lee also provides an account for indirect replies (Lee, '99) and *implicatures* (Lee & Wilks, '97) while assuming that interacting agents are rational and cooperative.

---

[9] In the TRINDIKIT implementation of the Dialogue Moves Engine the same mechanism have been adopted.

[10] Ascription is used in a restricted manner in combination with the planning algorithm, which considers only proposition within one belief space (that of the agent being simulated).

In our opinion, the work carried out by Lee can be further extended to several directions. For instance, a failure in ascribing the required knowledge to perform simulative planning may trigger the system's dialogue initiative. Moreover, ViewGen may serve as a base for implementing a *flexible Discourse Context Manager,* since planning recognition in ViewGen makes use of partial plans thus allowing the possibility of building incremental interpretation of dialogue acts.

### 3.4 Scenario description and reasoning

Inference is an essential component of the plan recognition process when derived information needs to be extracted from the factual knowledge. We agree with the pragmatic view than an intelligent dialogue system should incorporate a sufficient amount of background commonsense knowledge (Hobbs et al., '87) in order to capture the perlocutionary acts of expressed utterances. Reasoning about world and commonsense knowledge is required in plan recognition when at certain steps the action precondition can be satisfied just after few steps of inference or browsing the appropriate ontology. We use the *action language* FLP to encode this commonsense knowledge and perform practical reasoning on scenario descriptions.

### 3.5 Task-domain experts

We argue that the integration of the *capability brokering facility* will improve the portability of a general dialogue shell. We adopt the choice made in (Allen et al., '00) of having a clean separation between domain-independent and domain-specific information and processing. The use of an agent communication language supports a flexible and plug-and-play approach to system construction.

## 4 Conclusions

We have proposed a general agent-based framework for the modelling of an adaptive behaviour and its reproduction in artificial systems. We focused on NLP applications and in particular on the feasibility of the HERALD architecture for the design of dialogue systems.

We envision the design of a conversational shell for the COALA [12] system currently under development (Fatemi & Abou Khaled, '01). COALA is a digital audio-visual archive system with facilities for providing an effective content-based access. We plan to enrich the user interface by means of a dialogue interface that enables a more natural interaction for users with different levels of expertise.

We are also involved with the re-engineering of the GETA-RUN system (Delmonte, '92) GETA-RUN is a NLP system for the complete analysis of narrative texts. Although conceptually modular, it is implemented as an almost monolithic PROLOG program. We will approach the *agentification* of GETA_RUN as a reverse engineering problem. Further experimentations will deal with the design of novel analysis strategies based on the possibility offered by the HERALD architecture of having a rule-based control of the dynamics of computations. We will implement heuristics to dynamically change the dataflow and the linguistic processors' parameters. GETA-RUN agents will be reused as powerful linguistic components for future HERALD-based dialogue systems.

## References

(Allen et al., '00) Allen, J., Byron, D., Dzikovska, M., Ferguson, G., Galescu, L. and Stent, A., *Towards a Generic Dialogue Shell*, in *Natural Language Engineering*, 6(3), 1-16.

(Allen et al., '01) Allen, J., Ferguson, G. and Stent, A., *An architecture for more realistic conversational systems*. In *Proceedings of Intelligent User Interfaces 2001 (IUI-01)*, Santa Fe, NM,

(Ballim, '92) Ballim, A., *ViewFinder: A Framework for Representing, Ascribing and Maintaining Nested Beliefs of Interacting Agents,* University of Geneva, Geneve.

(Ballim, '93) Ballim, A., *Reasoning about Mental States: Formal Theories & Applications (Papers from the 1993 AAAI Spring Symposium)*, (Eds, Horty, J. and Shoham, Y.) AAAI press.

(Ballim et al., '00a) Ballim, A., Chappelier, J.-C., Pallotta, V. and Rajman, M., *ISIS: Interaction through Speech with Information Systems*. In *Proceedings of the 3rd International Workshop, TSD 2000*, Brno, Czech Republic, September 2000.

(Ballim et al., '00b) Ballim, A., Coray, G. and Pallotta, V., *A Hybrid approach to robust analysis of natural language data,* Swiss Federal Institute of Technology - Lausanne, Lausanne.

(Ballim & Pallotta, '99) Ballim, A. and Pallotta, V., *Proceedings of VEXTAL99 conference*, (Ed, Monte, R. D.), Venezia, 1999.

(Ballim & Pallotta, '00) Ballim, A. and Pallotta, V., *The role of robust semantic analysis in spoken language dialogue systems*. In *Proceedings of the third International Workshop on Human-Computer Conversation*, Bellagio, Italy, 3-5 July 2000.

(Ballim & Russell, '94) Ballim, A. and Russell, G., *LHIP: Extended DCGs for Configurable Robust Parsing*. In *Proceedings of the 15th International Conference on Computational Linguistics*, Kyoto, Japan,

(Ballim & Wilks, '90) Ballim, A. and Wilks, Y., *Proceedings of ECAI-90*, Stockholm, pp. 65--70.

---

[11] Stereotypical reasoning used in plan simulation may be influenced by information such as the Last Move or the Question Under Discussion from the Discourse Context.

[12] COALA stands for Content-Oriented Audiovisual Library Access.

(Ballim & Wilks, '91a) Ballim, A. and Wilks, Y., *Artificial Believers,* Lawrence Erlbaum Associates, 1991.

(Ballim & Wilks, '91b) Ballim, A. and Wilks, Y., *Beliefs, Stereotypes and Dynamic Agent Modelling*, in *User Modelling and User-Adapted Interaction,* 1**,** 33–65.

(Basili et al., '00) Basili, R., Mazzucchelli, M. and Pazienza, M. T., *An Adaptive and Distributed Framework for Advanced IR*. In *6th RIAO Conference (RIAO 2000)*, Paris,

(Bunt, '89) Bunt, H. C., *The structure of Multimodal Dialogue*, (Eds, Taylor, M. M., Neel, F. and Bouwhuis, D. G.) Elsevier Publishers.

(Chappelier et al., '99) Chappelier, J.-C., Rajman, M., Bouillon, P., Armstrong, S., Pallotta, V. and Ballim, A., *ISIS Project: final report,* Computer Science Department - Swiss Federal Institute of Technology, Lausanne.

(Cristea, '00) Cristea, D., *An Incremental Discourse Parser Architecture*. In *Second International Conference - Natural Language Processing - NLP 2000*, Patras, Greece,

(Cunningham et al., '95) Cunningham, H., Gaizauskas, R. and Wilks, Y., *A General Architecture for Text Engineering (GATE) -- a new Approach to Language Engineering R & D,* University of Sheffield, 1995.

(Delmonte, '92) Delmonte, R., *Linguistic and inferential processes in text analysis by computer,* Unipress, 1992.

(Fatemi & Abou Khaled, '01) Fatemi, N. and Abou Khaled, O., *Indexing and Retrieval of TV News Programs Based on MPEG-7*. In *Proceedings of the IEEE International Conference on Consumer Electronics (ICCE'2001)*, Los Angles, CA,

(Hobbs et al., '87) Hobbs, J., Croft, W., Davies, T. R., Edwards, D. and Laws, K. I., *Commonsense Metaphysics and Lexical Semantics*, in *Computational Linguistics,* 13(3-4)**,** 241-250.

(Hobbs et al., '93) Hobbs, J., Stickel, M., Appelt, D. and Martin, P., *Interpretation as Abduction*, in *Artificial Intelligence,* 63(1-2)**,** 69-142.

(Labrou & Finin, '97) Labrou, Y. and Finin, T., *A Proposal for a new KQML Specification* Computer Science and Electrical Engineering Department, University of Maryland Baltimore County, Baltimore, MD 21250.

(Larsson & Traum, '00) Larsson, S. and Traum, D., *Information state and dialogue management in the TRINDI Dialogue Move Engine Toolkit*, in *Natural Language Engineering,* 6**,** 323-340.

(Lee, '97) Lee, M., *Belief ascription in mixed initiative dialogue*. In *Proceedings of AAAI Spring Symposium on Mixed Initiative Interaction*,

(Lee, '98) Lee, M., *Belief, Rationality and Inference: A general theory of Computational Pragmatics,* University of Sheffield, Sheffield.

(Lee, '99) Lee, M., *Implicit goals in indirect replies*, in *Proceedings of the ESCA tutorial and research workshop on Interactive Dialogue in Multi-Modal Systems*, Kloster Irsee, Germany,

(Lee & Wilks, '96) Lee, M. and Wilks, Y., *An ascription-based approach to speech acts*. In *Proceedings of the 16th Conference on Computational Linguistics (COLING-96)*, Copenhagen,

(Lee & Wilks, '97) Lee, M. and Wilks, Y., *Eliminating deceptions and mistaken belief to infer conversational implicature*. In *Proceedings of the IJCAI-97 workshop on Conflict, Cooperation and Collaboration in Dialogue Systems*,

(Lieske & Ballim, '98) Lieske, C. and Ballim, A., *Rethinking Natural Language Processing with Prolog*, in *Proceedings of PAPPACTS98*, Practical Application Company, London, UK.

(McAllester & Rosenblitt, '91) McAllester, D. and Rosenblitt, D., *Systematic Nonlinear Planning*, in *Proceedings of the Ninth National Conference on Artificial Intelligence*,

(Nolke & Adam, '99) Nolke, H. and Adam, J.-M., (Eds.) (1999) *Approches modulaires: de la langue au discourse,* Delachaux et Niestlé, Lausanne.

(Pallotta, '99) Pallotta, V., *Reasoning about Fluents in Logic Programming*, in *Proceedings of the 8th International Workshop on Functional and Logic Programming*, Grenoble, FR,

(Pallotta, '00a) Pallotta, V., *Fluent Logic Programming Technical report,* EPFL, Lausanne.

(Pallotta, '00b) Pallotta, V., *A meta-logical semantics for Features and Fluents based on compositional operators over normal logic-programs*, in *First International Conference on Computational Logic*, London, UK,

(Poesio, '95) Poesio, M., *Disambiguation as (Defeasible) Reasoning about Underspecified Representations*, in *Papers from the Tenth Amsterdam Colloquium*, Amsterdam,

(Rao & Georgeff, '91) Rao, A. S. and Georgeff, M. P., *Proceedings of the 2nd International Conference on Principles of Knowledge Representation and Reasoning*, (Eds, Allen, J., Fikes, R. and Sandewall, E.) Morgan,, pp. 473--484.

(Shoham, '93) Shoham, Y., *Agent-Oriented Programming*, in *Artificial Intelligence,* 60**,** 51--92.

(Wickler, '99) Wickler, G. J., *Using Expressive and Flexible Action Representations to Reason about Capabilities for Intelligent Agent Cooperation* University of Edinburgh, Edinburgh.

(Wilks & Ballim, '87) Wilks, Y. and Ballim, A., *Proceedings of the 10th International Joint Conference on Artificial Inteligence*, Morgan Kaufmann,, pp. 118--124.